\pdfoutput=1

\documentclass{article}
\usepackage{arxiv}

\usepackage[utf8]{inputenc} 
\usepackage[T1]{fontenc}    
\usepackage{hyperref}       
\usepackage{url}            
\usepackage{booktabs}       
\usepackage{amsfonts}       
\usepackage{nicefrac}       
\usepackage{microtype}      
\usepackage{cleveref}       
\usepackage{lipsum}         
\usepackage{graphicx}
\usepackage{natbib}
\usepackage{doi}
\usepackage{lineno}
\usepackage{tikz}
\usepackage[colorinlistoftodos]{todonotes}        
\usepackage{soul}             
\usepackage{suffix}           
\usepackage{awesomebox}      
\usepackage{alertmessage}    
\usepackage{enumitem}
\usepackage{algorithm}
\usepackage{algpseudocode}
\usepackage[justification=centering]{caption}
\usepackage{multirow}

\newif\ifdraft
\newif\ifuniqueAffiliation

\ifdraft
  \linenumbers
\else
  \nolinenumbers
  \renewcommand\internallinenumbers{}
\fi
\usetikzlibrary{chains}

\hypersetup{
    pdftitle = {Embedding Attacks Project},
    pdfsubject = {cs.ML, cs.AI},
    pdfauthor = {Jiameng, Zafar},
    pdfkeywords = {MIA, PIA, embeddings},
}

\title{Embedding Attacks Project}
\renewcommand{\headeright}{Work Report}
\renewcommand{\undertitle}{Work Report}
\renewcommand{\shorttitle}{Embedding Attacks}

\date{}

\newcommand{\authorone}{Jiameng Pu}
\newcommand{\affilone}{PnS Privacy Innovation Lab, TikTok}
\newcommand{\emailone}{\texttt{jiameng.pu@tiktok.com}}

\newcommand{\authortwo}{Zafar Takhirov}
\newcommand{\affiltwo}{PnS Privacy Innovation Lab, TikTok}
\newcommand{\emailtwo}{\texttt{zafar.takhirov@tiktok.com}}

\ifuniqueAffiliation 
  \author{
    \authorone \\
    \affilone \\
    \emailone \\
    \And
    \authortwo \\
    \affiltwo \\
    \emailtwo \\
  }
\else
  \usepackage{authblk}
  
  \author{%
    \authorone\thanks{\emailone}%
  }
  \author{%
    \authortwo\thanks{\emailtwo}%
  }
  \affil{\affilone}
\fi

\newcommand\numberbox[3][abnote]{\awesomebox[#1]{\aweboxrulewidth}{#2}{#1}{#3}}

\begin{document}
\maketitle


\begin{abstract}

This report summarizes all the MIA experiments (Membership Inference Attacks) of the Embedding Attack Project, including threat models, experimental setup, experimental results, findings and discussion.
Current results cover the evaluation of two main MIA strategies (loss-based and embedding-based MIAs) on 6 AI models ranging from Computer Vision to Language Modelling.
There are two ongoing experiments on MIA defense and neighborhood-comparison embedding attacks.
Results will be added to this report once it is done.

\end{abstract}



\section{Introduction}\label{sec:introduction}

In recent years, machine learning models have become integral in various applications, from computer vision to natural language processing.
Yet, deploying these models poses privacy concerns, as they may inadvertently reveal sensitive information embedded in the training data.
This report focuses on the hidden or intermediate layer outputs (i.e., embeddings) of ML models, investigating the potential for sensitive information leakage from these embeddings.

\paragraph{Defining Embedding Layers and Embeddings}
An embedding layer, a type of hidden layer in neural networks, transforms input information from a high-dimensional space to a lower-dimensional one.
This transformation aids in elucidating relationships among inputs and enhances data processing efficiency.
We refer to hidden or intermediate layers in neural networks as embedding layers, and their outputs as embeddings.

\paragraph{The Importance of Investigating Privacy Leakage in Embedding Layers}
Investigating privacy leakage in embedding layers is crucial for several reasons.
First, as embeddings could facilitate data transfer, understanding their privacy implications is vital, especially given the stringent privacy regulations governing cross-border data transfers.
Second, embedding layers, such as those in federated learning systems or split learning systems, are often exposed and vulnerable to privacy attacks, enabling attackers to infer sensitive information.

\paragraph{Privacy Perspectives in this Investigation}
Our investigation focuses on embedding-based privacy attacks against machine learning models to assess potential privacy risks in their embedding layers.
Specifically, we explore the feasibility of inferring membership and property information of data samples from their embeddings, addressing a significant privacy concern.

\paragraph{Real-World Implications}
This research has several practical implications:
\begin{itemize}
\item In cross-border data/model transfers, subject to strict legislation, we aim to assess the risks associated with transferring embeddings for downstream training tasks.
\item Investigating embedding-based attacks in recommender systems could inform the development of more privacy-preserving models.
\item In e-commerce, understanding how embeddings might reveal customer preferences and habits is crucial for customer data protection.
\item The potential for inferring sensitive information from text data, even when encoded as embeddings, highlights privacy concerns in NLP applications.
\end{itemize}

\subsection*{Objectives and Tasks}

The primary objective of this study is to measure privacy leakage in the embedding layers of ML models.
Our main tasks are summarized as follows:
\begin{enumerate}
    \item \textbf{Examination of Privacy Leakage Across Models:} We investigate privacy leakage from embedding layers in six classification models, spanning domains from Computer Vision to Language Modeling.

    \item \textbf{Types of Privacy Attacks Analyzed:}
        \begin{enumerate}[label=\alph*.]
            \item \textbf{Membership Inference Attacks (MIAs):} MIAs aim to determine whether a specific data sample was part of the model's training set.
            We focus on how embedding-based MIAs can exploit this membership information.
            
            \item \textbf{Property Inference Attacks (PIAs):} These attacks analyze how embedding-based PIAs can infer properties or attributes of data samples.
        \end{enumerate}

    \item \textbf{Assumptions and Attack Strategies:} We assume threat models that provide attackers with varying levels of resource leverage (detailed in Section 3).
    We implement several attack strategies (described in Section 4.2) to measure the degree of privacy leakage under different model settings.
    
    \item \textbf{In-depth Analysis of MIA Strategies:} We particularly focus on designing different strategies for MIAs, exploring:
        \begin{enumerate}[label=\alph*.]
            \item the impact of embedding layer depth on the severity of privacy leakage,
            \item the influence of model overfitting on attack performance,
            \item the extent of privacy compromise due to label leakage.
        \end{enumerate}
\end{enumerate}



\section{Threat Models}\label{sec:threat-models}

\begin{figure}
    \centering
    \includegraphics[width=\textwidth]{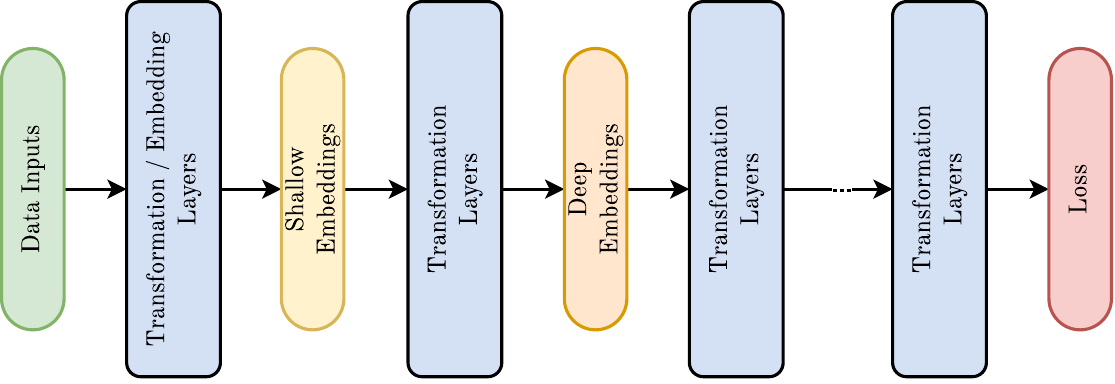}
    \caption{Demonstration of embeddings from shallow to deep within an ML model.}
    \label{fig:threat-embeddings-shallow-deep}
\end{figure}

\subsection{Membership Inference Attack}\label{sec:threat-models:mia}

\paragraph{Targeted Models}
We analyze five classification models trained on various data modalities.
These models are categorized as follows:
\begin{enumerate}
    \item \textbf{Image Classification Models:}
        \begin{itemize}
            \item Exploration Case \#1: CNN Trained on CIFAR-10
            \item Exploration Case \#2: ViT Trained on Snacks dataset
        \end{itemize}
    \item \textbf{MLP Classification Models:}
        \begin{itemize}
            \item Exploration Case \#3: MLP Classifier Trained on PurChase100
        \end{itemize}
    \item \textbf{Text Classification Models:}
        \begin{itemize}
            \item Exploration Case \#4: BERT-based Sentiment Classifier Trained on IMDb Reviews
            \item Exploration Case \#5: Encoder-only Transformer Classifier Trained on AGNews
        \end{itemize}
\end{enumerate}

\paragraph{Attacker's Capabilities and Goals}
To evaluate privacy leakage from embedding layers, we consider scenarios where attackers have moderate data resource leverage.
We avoid overly complex settings, like requiring attackers to create shallow datasets for training models.
Such settings are realistic, considering potential internal threats in organizations due to lax data management.
They also help estimate the maximum potential of attackers, providing insights into worst-case privacy leakage scenarios.

We focus on membership inference attacks (MIAs) that determine a data sample's membership based on its embeddings, i.e., latent representations from a specific hidden layer of an ML model.
We assume the attacker can access a limited set of embeddings from both training (members) and non-training samples (non-members).
The attacker's goal is to determine the membership status of a new data sample – whether it belongs to the model's original training set.

Key assumptions include:
\begin{itemize}
    \item Access to the model
    \item Familiarity with the model's architecture
\end{itemize}


\subsection{Property Inference Attack}\label{sec:threat-models:pia}

\paragraph{Targeted Models}
For property inference attacks, we utilize a CNN classifier trained on CelebA, a large-scale face attributes dataset with over 200K celebrity images, each annotated with 40 attributes.
\begin{itemize}
    \item Exploration Case \#6: CNN Trained on CelebA
\end{itemize}

\paragraph{Attacker's Capabilities and Goals}
In property inference attacks, we focus on a CNN classifier trained to identify smiling faces.
Similar to MIA, we assume the attacker has access to embeddings from the model, which are latent representations from specific hidden layers.

The attacker aims to infer secondary attributes, such as gender and facial characteristics, unrelated to the primary classification task (smiling or not).
They can access a small auxiliary dataset of human-face images, which they can manually label for properties they aim to infer.
The attack leverages embeddings computed by the original CNN classifier.

Key assumptions include:
\begin{itemize}
    \item Access to the model
    \item Knowledge of the model's architecture
\end{itemize}



\section{Experimental Setup}\label{sec:experimental-setup}

\subsection{Models and Datasets}\label{sec:experimental-setup:models-datasets}

\paragraph{Image Classification Models}
For image classification, we evaluated three models: two CNNs and one Vision Transformer (ViT).
The first CNN was trained on the CIFAR-10 dataset, which includes 60,000 color images in ten classes, split evenly between 50,000 training and 10,000 test images.
The model architecture consists of 3 Conv2D layers, 3 MaxPool2D layers, 3 BatchNormalization layers, and 2 fully connected layers, employing ReLu activation and trained over 50 epochs.
The second CNN was trained on the CelebA dataset, a large-scale face attributes dataset with over 200K celebrity images, each annotated with 40 attributes.
This model follows a similar architecture to the first, with adjustments for the CelebA dataset's specifics.
The third model, a ViT, was trained on the Snacks dataset, derived from Google Open Images, containing 6745 images of 20 snack types.
The ViT, pretrained on ImageNet-21k, processes images as fixed-size patches and includes a [CLS] token for classification tasks.

\begin{table}
  \centering
  \caption{Training and testing performances of image classification models}
  \label{tab:models-datasets:image}
  \begin{tabular}{llll}
    \toprule
    Model & Dataset       & Training Accuracy & Validation Accuracy \\
    \midrule
    CNN   & CIFAR-10      & 1.00              & 0.76                \\
    CNN   & CelebA        & 0.88              & 0.87                \\
    ViT   & Snacks        & 0.98              & 0.81                \\
    \bottomrule
  \end{tabular}
\end{table}

\paragraph{Multilayer Perceptron Classifiers}
In the realm of tabular data, we used a Multilayer Perceptron (MLP) Classifier trained on the Purchase100 dataset.
This dataset comprises 100,000 anonymized purchase records across various product categories.
The MLP model consists of a 4-layer fully connected neural network with layers [1024, 512, 256, 100] and Tanh activation functions, trained for 50 epochs.

\begin{table}
  \centering
  \caption{Training and testing performances on tabular data}
  \label{tab:models-datasets:tabular}
  \begin{tabular}{llll}
    \toprule
    Model & Dataset     & Training Accuracy & Validation Accuracy \\
    \midrule
    MLP   & Purchase100 & 0.98              & 0.85                \\
    \bottomrule
  \end{tabular}
\end{table}

\paragraph{NLP Text Classifiers}
For text classification, we employed two models: a BERT-based Sentiment Classifier and an Encoder-only Transformer.
The BERT-based classifier was adapted for the IMDb Movie Reviews dataset, comprising 50,000 reviews evenly divided between positive and negative sentiments.
A Dropout layer and a Linear layer were added on top of the pretrained BERT model.
The Encoder-only Transformer, with an architecture borrowed from XXX, was trained on the AG News dataset, consisting of news articles across four topics.

\begin{table}
  \centering
  \caption{Training and testing performances on language data}
  \label{tab:models-datasets:nlp}
  \begin{tabular}{llll}
    \toprule
    Model       & Dataset & Training Accuracy & Validation Accuracy \\
    \midrule
    BERT        & IMDB    & 0.99              & 0.89                \\
    Transformer & AGNews  & 1.00              & 0.93                \\
    \bottomrule
  \end{tabular}
\end{table}


\subsection{Attack Methodologies}\label{sec:experimental-setup:attack-methodologies}

\subsubsection{Membership Inference Attack}\label{sec:attack-methodologies:mia}
Membership inference attacks (MIAs) aim to determine whether a given data record was part of a model's training dataset.
This is a critical concern as learning that a record was used to train a model indicates information leakage, potentially leading to a privacy breach.

\paragraph{Attack Methodology}
Neural-Network-based MIAs (NN MIAs) extract features from the target model to train an NN-based binary classifier.
This classifier determines whether a sample belongs to the training dataset.
The features can be either intermediate-layer outputs (i.e., latent-space embeddings) or final layer outputs, such as prediction loss and prediction vectors.
Our focus is on understanding the effectiveness of inferring membership based on latent-space embeddings.
Therefore, embeddings of data samples are extracted as features to train the attack model.
Additionally, we include two baseline settings using prediction loss and vectors as training features.
Consequently, we have three attack settings:

\begin{enumerate}
    \item \textit{Embedding-based NN MIA (Attack Setting \#1)} uses embeddings from intermediate layers of deep learning models.
    A Multilayer Perceptron is trained on these embeddings from a balanced set of members and non-members.
    \item \textit{Prediction-based NN MIA (Attack Setting \#2)} utilizes prediction vectors, which typically consist of probabilities or scores for each class.
    A Multilayer Perceptron is trained on these vectors.
    \item \textit{Loss-based NN MIA (Attack Setting \#3)} employs prediction loss to quantify the alignment of model predictions with actual values.
    A Multilayer Perceptron is trained on this prediction loss.
\end{enumerate}

\begin{figure}
    \centering
    \includegraphics[width=\textwidth]{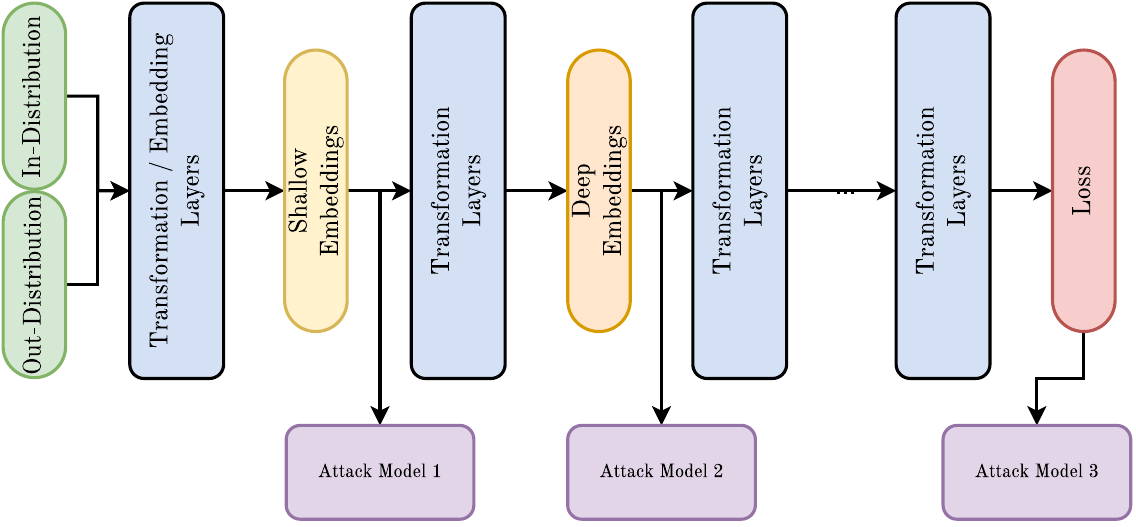}
    \caption{The attack strategy of three neural-network-based MIAs}
    \label{fig:nn-based-mia}
\end{figure}

\subsubsection{Property Inference Attack}\label{sec:attack-methodologies:pia}
Property inference attacks (PIAs) infer specific properties of an image sample, such as gender, based on its embeddings.
These inferred properties are unrelated to the model's original classification task (e.g., determining if a person is smiling).

\paragraph{Attack Methodology}
Neural-Network-based PIAs (NN PIAs) employ a Multilayer Perceptron as the architecture for the attack model.
The attacker uses an auxiliary dataset labeled with the properties they aim to infer.
If such a dataset is not available, the attacker can scrape images from the Internet and manually label them.
The data samples from this auxiliary dataset are fed into the model to obtain their embeddings, which are then used to train the attack model along with the property labels.
A separate MLP attack model is trained for each property of interest.

\begin{figure}
    \centering
    \includegraphics[width=\textwidth]{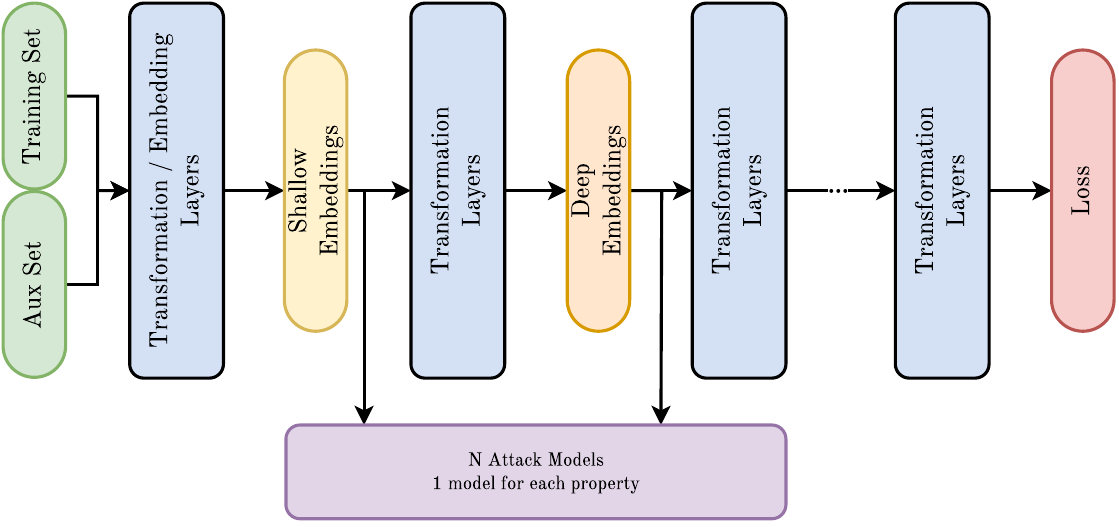}
    \caption{The attack strategy of neural-network-based PIAs}
    \label{fig:nn-based-pia}
\end{figure}



\section{Experimental Results}\label{sec:experimental-results}

\subsection{Key Findings}\label{sec:experimental-results:key-findings}

\numberbox{F1}{\internallinenumbers{}
The degree of overfitting in models directly affects their vulnerability to Membership Inference Attacks (MIAs).
Avoiding model overfitting is a crucial step in defending against MIAs.}

\paragraph{What is overfitting?}
Overfitting occurs when a model fits the training data well but fails to generalize to new, unseen data.
This happens when the model learns patterns specific to the training data that do not apply to other datasets.
Overfitting is typically identified by monitoring validation metrics, such as loss or accuracy.
These metrics often stop improving after a certain number of training epochs and may even begin to worsen, while the training metrics continue to improve.

\paragraph{Which models are most prone to overfitting?}
Nonparametric and nonlinear models, with their greater flexibility in learning target functions, are more susceptible to overfitting.
Many machine learning algorithms limit the amount of detail learned to prevent overfitting.
We investigated two cases:

\paragraph{Setting \#1: Non-overfitted ViT model vs. Overfitted ViT model}

We trained the Vision Transformer (ViT) model on the Snacks dataset (Exploration Case \#2) to obtain two versions with varying degrees of overfitting, indicated by the gap between training and testing performances.

\begin{table}
  \centering
  \caption{ViT model performances on Snacks dataset and NN MIA performances under two different model settings}
  \label{tab:findings:vit}
  \begin{tabular}{llllll}
    \toprule
      & Overfitting & Training  & Testing   & Loss-based    & Embedding-based \\
      & Degree      & Accuracy  & Accuracy  & NN MIA (AUC)  & NN MIA (AUC)    \\
    \midrule
    Model 1 & Little  & 0.95  & 0.89  & 0.63  & 0.52  \\
    Model 2 & Strong  & 0.99  & 0.83  & 0.65  & 0.53  \\
    \bottomrule
  \end{tabular}
\end{table}

\paragraph{Setting \#2: Overfitted model trained on Purchase100 and Non-overfitted model trained on Purchase2.}
We trained two MLP classifiers on the Purchase dataset.
The first MLP, trained on Purchase100, was prone to overfitting due to its complex decision boundary.
Conversely, the second MLP, trained on Purchase2, converged quickly and showed similar training and testing performance (99\% and 98\% accuracy, respectively), indicating less overfitting.
Table \ref{tab:findings:mlp} presents these models' performances and the corresponding MIA performances, highlighting the stark differences between overfitted and non-overfitted models.

\begin{table}
  \centering
  \caption{MLP model performances on Purchase dataset and NN MIA performances under two different settings}
  \label{tab:findings:mlp}
  \begin{tabular}{lccccc}
    \toprule
    Dataset & Overfitting & Training  & Testing   & Loss-based    & Embedding-based \\
            & Degree      & Accuracy  & Accuracy  & NN MIA (AUC)  & NN MIA (AUC)    \\
    \midrule
    PurChase100 & Little & 0.99 & 0.98 & 0.65 & 0.55 \\
    PurChase2   & Strong & 0.99 & 0.89 & 0.51 & 0.46 \\
    \bottomrule
  \end{tabular}
\end{table}

\numberbox{F2}{\internallinenumbers{}
Despite embeddings containing rich information, disentangling membership information from semantic information is extremely challenging.
Loss-based NN MIAs are generally more effective than embedding-based NN MIAs across both vision and text classifiers.}

We conducted loss-based and embedding-based NN MIAs on five different models described in Section 4.1.
The results show that loss-based NN MIAs generally outperform embedding-based NN MIAs on all models except for the BERT sentiment classifier.
For instance, the loss-based NN MIA achieved up to 78\% AUC on the CNN model trained with CIFAR-10, whereas the embedding-based NN MIA only reached 55\% AUC.
On the MLP model trained with PurChase100, the loss-based NN MIA attained 65\% AUC, compared to 53\% AUC for the embedding-based approach.
A notable exception was observed with the ViT model trained on Snacks, where the embedding-based NN MIA performed below random chance, achieving only 46\% AUC.

\begin{table}
  \centering
  \caption{Attack performances of loss-based NN MIA and embedding-based NN MIA on five different models}
  \label{tab:findings:attacks}
  \begin{tabular}{llllll}
    \toprule
      & Training & Testing & Loss-based & Prediction based & Embedding-based \\
      & Accuracy & Accuracy & NN MIA (AUC) & NN MIAs (AUC) & NN MIA (AUC) \\
    \midrule
    MLP / PurChase100 & 0.99 & 0.89 & 0.65 & ?.?? & 0.55 \\
    CNN / CIFAR10 & 1.00 & 0.77 & 0.78 & 0.73 & 0.55 \\
    ViT / Snacks & 0.98 & 0.81 & 0.65 & 0.55 & 0.53 \\
    BERT / IDMb & 0.99 & 0.89 & 0.59 & 0.56 & 0.61 \\
    Transformer-encoder & 1.00 & 0.93 & 0.56 & 0.54 & 0.53 \\
    \bottomrule
  \end{tabular}
\end{table}

\numberbox{F3}{\internallinenumbers{}
Embeddings from deeper layers of a model's hierarchy reveal more membership information.}

We demonstrated this point using MLP trained on PurChase100 and CNN trained on CIFAR-10.
For both models, we performed embedding-based NN MIAs using embeddings from layers at varying depths.
Deeper layers, closer to the model's final output, provided more revealing embeddings.
Table \ref{tab:findings:depth} shows the increasing effectiveness of attacks using embeddings from deeper layers, suggesting that these layers tend to reveal more membership information.

\begin{table}
  \centering
  \caption{Performance of Embedding-based NN MIA based on embeddings from different depths}
  \label{tab:findings:depth}
  \begin{tabular}{lccccc}
    \toprule
      & Training & Testing & Loss-based & Shallow Embedding & Deep Embedding \\
      & Accuracy & Accuracy & NN MIA (AUC) & \multicolumn{2}{c}{NN MIA (AUC)} \\
    \midrule
    MLP / PurChase100 & 1.00 & 0.98 & 0.65 & 0.52 & 0.55 \\
    CNN / CIFAR10 & 0.99 & 0.77 & 0.78 & 0.51 & 0.55 \\
    \bottomrule
  \end{tabular}
\end{table}

\awesomebox[white][]{0pt}{\faBookmark[regular]}{black}{\internallinenumbers{}
We propose that membership information accumulates in deeper layers of classification models.
In initial layers, embeddings intertwine semantic and membership information, making them difficult to separate effectively.
However, as layers progress, semantic information in embeddings reduces, while membership information becomes more concentrated.
This concentration is maximal at the last layer, explaining the superior performance of attacks using deeper embeddings and the consistent effectiveness of loss-based NN MIAs.
To effectively separate membership from semantic information, more sophisticated attack strategies are necessary.
We discuss strategies for stronger attacks in the section on Ongoing Experimental Threads.}

\numberbox{F4}{\internallinenumbers{}
Membership information is significantly compromised if embeddings and corresponding labels are provided together.}

Embedding-based NN MIAs generally underperform compared to loss-based NN MIAs when attackers lack access to ground-truth labels, a common scenario in split learning and federated learning.
However, this scenario changes if embeddings and their labels are shared or stolen for training downstream tasks.
Such a transfer of embeddings and labels, despite strict data transfer regulations, necessitates investigating the severity of privacy leakage under these circumstances.
We trained a shadow MLP for each model setting described in Section 4.1 and repeated the loss-based NN MIA using losses from the shadow MLP.
The results, shown in table \ref{tab:findings:ground-truth}, reveal that this strategy outperforms embedding-based NN MIAs but falls short compared to direct loss-based NN MIAs due to simulation inaccuracies.

\begin{table}
    \centering
    \caption{Attack performances when both embeddings and ground-truth labels are available}
    \label{tab:findings:ground-truth}
    \begin{tabular}{lccccc}
        \toprule
        ~ & \multicolumn{2}{c}{Accuracy} & \multicolumn{3}{c}{NN MIA (AUC)} \\
        \cmidrule(r){2-3} \cmidrule(r){4-6}
        ~ & Training & Testing & Loss-based & Embedding-based & With Ground-truth Label \\
        \midrule
        MLP / PurChase100 & 0.99 & 0.89 & 0.65 & 0.55 & 0.66 \\
        CNN / CIFAR10 & 1.00 & 0.77 & 0.78 & 0.55 & 0.69 \\
        ViT / SNACK & 0.98 & 0.83 & 0.65 & 0.53 & 0.68 \\
        BERT / IMdB & 0.99 & 0.89 & 0.59 & 0.61 & 0.61 \\
        Transformer-encoder & 1.00 & 0.93 & 0.56 & 0.53 & 0.53 \\
        \bottomrule
    \end{tabular}
\end{table}

\numberbox{F5}{\internallinenumbers{}
Even without ground-truth labels, attackers can infer pseudo-labels through unsupervised clustering algorithms or label inference attacks, potentially compromising membership information.}

Attackers with access only to embeddings might still infer labels using methods like unsupervised clustering or label inference attacks.
We assume attackers employ unsupervised clustering algorithms, like K-means, to group embeddings and assign pseudo-labels.
These pseudo-labels are then used to train a shadow model simulating the loss distribution.
Our experiments showed that this strategy could achieve a 59\% AUC score against the MLP model trained on Purchase100.
However, its effectiveness is limited when embeddings are challenging to cluster accurately, as seen in the case of the CNN trained on CIFAR-10.

\begin{table}
    \centering
    \caption{Attack performances with embeddings and pseudo-labels estimated via unsupervised clustering}
    \label{tab:findings:pseudo-labels}
    \begin{tabular}{lcccccc}
        \toprule
        ~ & \multicolumn{2}{c}{Accuracy} & \multicolumn{4}{c}{NN MIA (AUC)}  \\
        \cmidrule(r){2-3} \cmidrule(r){4-7}
        ~ & Training & Testing & Loss-based & Emb.-based    & + Ground-truth    & + Pseudo-labels \\
        ~ & ~ & ~ & ~ & ~                                   & Labels            & + clustering \\
        \midrule
        MLP / PurChase100 & 0.99 & 0.89 & 0.65 & 0.55 & 0.66 & 0.59 \\
        CNN / CIFAR10 & 1.00 & 0.77 & 0.98 & 0.55 & 0.69 & 0.54 \\
        \bottomrule
    \end{tabular}
\end{table}

\paragraph{Other possible attack strategy (i.e., Label Inference Attack).}
Label inference attacks are feasible in federated learning or split learning settings.
In these scenarios, sensitive information can be leaked through communicated gradients or model parameters, even when raw data isn't shared.
For example, in horizontal Federated Learning, techniques have been demonstrated that allow an honest-but-curious server to uncover raw features and labels of a device's data, based on the model architecture, parameters, and communicated gradient.
Further, Zhao et al. have shown that ground truth labels of examples can be extracted by exploiting the directions of gradients connected to different class logits.

\numberbox{F6}{\internallinenumbers{}
Despite embeddings-based NN MIAs achieving less than 60\% AUC for all evaluated models, embeddings are not inherently resistant to privacy leakage, as their rich semantic information is susceptible to property inference attacks.}

Our focus on membership leakage revealed that deeper embeddings tend to leak more membership information, but this does not imply that shallow embeddings are safe from other privacy perspectives.
Since embeddings are latent-space representations rich in semantic features, they are vulnerable to property inference attacks, particularly in shallower layers.
For instance, we trained a CNN classifier on CelebA to distinguish whether a person's mouth is open or not.
Despite its training specificity, embeddings from this model could still leak several facial attributes.
An attacker could gather an auxiliary dataset with face images and corresponding labels for targeted attributes and use these to train an attack model.
Our experiments showed that embeddings from shallower layers leaked more facial attributes, as indicated by higher attack performances, while deeper embeddings were less revealing.

\begin{table}
    \centering
    \caption{NN PIA performance on training and testing sets of the CNN classifier trained on the CelebA dataset with the Mouth Slightly Open (MSO) target.
    The model performance on MSO is measured at 0.87 and 0.85 for training and testing sets.}
    \label{tab:findings:pia-mso}
    \begin{tabular}{lcccc}
        \toprule
        ~ & \multicolumn{2}{c}{Training Set} & \multicolumn{2}{c}{Test Set} \\
        \cmidrule(r){2-3} \cmidrule(r){4-5}
        ~ & Shallow embeddings & Deep embeddings & Shallow embeddings & Deep embeddings \\
        \midrule
        Male & 0.827 & 0.832 & 0.723 & 0.729 \\
        High\_Cheekbones & 0.735 & 0.696 & 0.745 & 0.688 \\
        Smiling & 0.844 & 0.833 & 0.826 & 0.807 \\
        Wearing\_Lipstick & 0.818 & 0.773 & 0.732 & 0.634 \\
        \bottomrule
    \end{tabular}
\end{table}


\subsection{Neighborhood-Based Embeddings}\label{sec:experimental-results:neighborhood-embeddings}
\awesomebox[white][]{0pt}{\faBookmark[regular]}{black}{\internallinenumbers{}
In our experiments, vision models have shown greater susceptibility to Membership Inference Attacks (MIAs) compared to NLP models, with both loss-based (AUC 78\%) and embedding-based MIAs (AUC 69\%) being more effective.
Transformer-based NLP models, in particular, demonstrate higher resistance to loss-based MIA.
This necessitates the development of stronger MIAs, especially for attacking text models.
Currently, we are implementing a neighborhood-comparison-based attack method for both vision and NLP models.}

Our previous experiments indicate that while embedding-based NN MIAs can be effective against some models, they generally underperform compared to loss-based NN MIAs, as detailed in Findings 2 and 3.
Moreover, irrespective of the attack method, MIAs tend to achieve lower performance on NLP models.
An exception was observed with the BERT sentiment classifier, where our attack achieved an AUC score close to 60\%.
This discrepancy could be related to the differing levels of overfitting in the NLP models we evaluated; the smaller gap between training and testing performance suggests these models are less prone to overfitting compared to the vision classification models we examined.

To develop more potent MIAs, we are currently implementing a neighborhood comparison-based attack, inspired by the approach proposed by \cite{mattern2023membership}.
This paper introduces neighborhood attacks that compare model scores for a given sample against scores of synthetically generated neighbor samples, thereby circumventing the need for access to the training data distribution.
We plan to adapt this attack to focus on embedding distance comparison.


\subsection{Defenses}\label{sec:experimental-results:defenses}

The results in this report indicate that the transfer of pretrained models, even partially, opens the possibility of membership inference.
For example, the simplest attack on the each layer of a model could be defined as an objective
\[
\min_{\hat{x}\in\mathcal{X}(\mathcal{V})} ||\Phi_d(\hat{x})-\Phi_d(x^*)||^2_2
\]
where $\Phi_d$ is the targeted model up to layer $d$, $\Phi_d(x^*)$ is the output of the dataset, and $\hat{x}$ is the input that needs to be reverse engineered.
In such a simple setup, without the presence of any defenses, a potential adversary can perfectly reconstruct the inputs, assuming white-box model.

More realistically, consider algorithm \ref{alg:whitebox:generic} for the inversion.
In that setup, given some inversion optimization $\mathcal{O}$, it is possible to reverse engineer a set of outputs to infer the input that was used.

\begin{algorithm}
    \caption{Whitebox inversion attack}\label{alg:whitebox:generic}
    \begin{algorithmic}
        \Require Whitebox model $\Phi$, Lower layer representation $\Psi$
        \Require Target $y=\Phi(x^*)$, Auxiliary dataset $D_{aux}$, Vocabulary $w_i \in W$
        \Require Inversion optimization objective $\mathcal{O}$
        \State $D_{train} \gets \{\Phi(x_i), \Psi(x_i)\} \forall x_i \in D_{aux}$
        \State $M \gets \min_M\left(||M(\Phi(x_i))-\Psi(x_i)||^2_2\right) \forall x_i \in D_{aux}$ \Comment{Mapping from result to lower layer representation}
        \State $z \gets \min_z\left(\mathcal{O} + \lambda\|z\|_1\right)$
        \State \Return $\hat{x} = \{w_i | i=\arg\max_{j=\{1,\dots,l\}}{z_j}\}$ \Comment{Or any other inference rule}
    \end{algorithmic}
\end{algorithm}

\begin{figure}
    \centering
    \includegraphics[width=\textwidth/2]{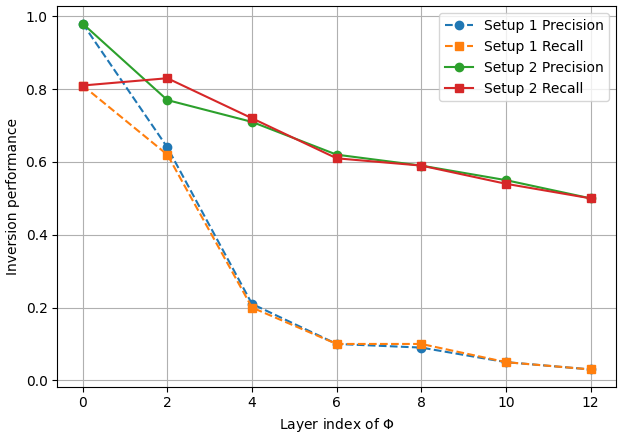}
    \caption{Inverting lower layer representations and embedding of BERT.}
    \label{fig:defenses:bert}
\end{figure}

Figure \ref{fig:defenses:bert} shows the attack performance on BERT model using two different inversion optimization techniques.
In that experiment two setups were evaluated:

\textbf{Setup 1}
    \begin{itemize}
        \item Inversion objective $\min_z{||\Phi\left(\mathbb{V}^T \cdot \mbox{softmax}(z/T)\right) - M(\Phi(x^*))||_2^2}$
        \item Inference rule $\hat{x} = \{w_i|i=\arg\max z_j\}_{j=1}^l$
        \item Temperature $T = 0.05$
        \item Optimizer `Adam` with $lr = 0.001$
    \end{itemize}
\textbf{Setup 2}
    \begin{itemize}
        \item Inversion objective $\min_{z\in\mathbb{R}^{|\mathcal{V}|}_{0+}}{||\mathbb{V}^T \cdot z - M(\Phi(x^*))||_2^2 + \lambda||z||_1}$
        \item Inference rule $\hat{x} = w_i \mathbb{1}_{z_i > \tau}$
        \item L1 penalty on sparsity: $\lambda=0.1$
        \item Sparsity threshold: $\tau = 0.01$
        \item Optimizer: `Adam` with $lr = 0.001$
    \end{itemize}

Notice that the inversion performance gets better as attacker gains access to the lower level representation.
Table \ref{tab:findings:defenses} shows some of the best results that were acquired by running whitebox and blackbox attacks on the last layers of several models.
In that table `Custom` model corresponds to a 4 layer MLP/linear head attached to BERT's first transformer layer.
Our goal was to evaluate the performance of an attacker in case of a partial model transfer.

\begin{table}[]
\centering
    \caption{Results of inversion attacks and simple defenses on whitebox and blackbox models. Defense is performed using noisy self-distillation.}
    \label{tab:findings:defenses}
    \begin{tabular}{llllll}
        \toprule
        ~ & ~ & \multicolumn{2}{c}{No Defense} & \multicolumn{2}{c}{With Defense} \\
        Threat Model & Target Model & Precision & Recall \\
        \midrule
        \multirow{3}{*}{White Box} & LSTM & 0.74 & 0.78 & 0.64 & 0.57 \\
         & Custom & 0.65 & 0.54 & 0.65 & 0.54 \\
         & BERT & 0.67 & 0.73 & 0.5 & 0.49 \\
         ~\\
        \multirow{3}{*}{Black Box} & LSTM & 0.91 & 0.39 & 0.59 & 0.51 \\
         & Custom & 0.81 & 0.26 & 0.53 & 0.60 \\
         & BERT & 0.85 & 0.36 & 0.62 & 0.49 \\
         \bottomrule
    \end{tabular}
\end{table}

\paragraph{Defense} In this setup the defense that was explored was ``Noisy Self-Distillation'' based on \cite{zhang2019teacher}.
The main difference is that during training we injected noise into the soft-labels.
Our goal was not to guarantee privacy, but rather evaluate the performance of a potential attacker.
Hence, we claim no privacy guarantees.
However, one cannot help by draw parallels between noisy self-distillation and differential privacy (see \cite{Abadi_2016} as an example).



\section{Conclusion}\label{sec:conclusion}

In this study, we have explored embedding-based membership inference attacks (MIAs) and property inference attacks (PIAs), employing neural network-based models to deduce the membership status or properties of data samples.
Our investigations across six machine learning classification models, spanning domains from Computer Vision to Language Modeling, have led to six key findings.

Firstly, the level of overfitting in targeted models significantly influences their vulnerability to MIAs. Models that exhibit higher degrees of overfitting are more susceptible to these attacks.
Secondly, the depth of hidden layers from which embeddings are extracted directly impacts the effectiveness of MIAs. 
We find that disentangling membership information from semantic content in embeddings is challenging, making loss-based neural network attacks generally more effective than embedding-based ones across both vision and text classifiers.

Furthermore, the presence of ground-truth labels can substantially increase the risk of membership information leakage in embeddings.
We demonstrate the feasibility of approximating labels using unsupervised clustering algorithms like K-means, highlighting another potential avenue for attack.
Lastly, our additional experiments confirm that the rich semantic information contained within embeddings makes them prone to property inference attacks.

These findings emphasize the need for careful consideration of model overfitting and the depth of embedding layers when assessing the privacy risks associated with deploying machine learning models, particularly in scenarios where embeddings and labels may be exposed.


\bibliographystyle{apalike}
\bibliography{ms}

\begin{thebibliography}{}

\bibitem[Abadi et~al., 2016]{Abadi_2016}
Abadi, M., Chu, A., Goodfellow, I., McMahan, H.~B., Mironov, I., Talwar, K., and Zhang, L. (2016).
\newblock Deep learning with differential privacy.
\newblock In {\em Proceedings of the 2016 ACM SIGSAC Conference on Computer and Communications Security}, CCS’16. ACM.

\bibitem[Mattern et~al., 2023]{mattern2023membership}
Mattern, J., Mireshghallah, F., Jin, Z., Schölkopf, B., Sachan, M., and Berg-Kirkpatrick, T. (2023).
\newblock Membership inference attacks against language models via neighbourhood comparison.

\bibitem[Zhang et~al., 2019]{zhang2019teacher}
Zhang, L., Song, J., Gao, A., Chen, J., Bao, C., and Ma, K. (2019).
\newblock Be your own teacher: Improve the performance of convolutional neural networks via self distillation.

\end{thebibliography}

\clearpage

\end{document}